\documentclass[journal]{IEEEtran}
\usepackage{amsmath,amsfonts}
\usepackage{algorithmic}
\usepackage{algorithm}
\usepackage{array}
\usepackage{textcomp}
\usepackage{stfloats}
\usepackage{url}
\usepackage{verbatim}
\usepackage{cite}
\usepackage{graphicx}
\usepackage{multirow}
\usepackage{booktabs}
\usepackage{xcolor}
\usepackage{adjustbox}

\usepackage{hyperref}
\hypersetup{
    colorlinks=true,
    citecolor=blue,
    linkcolor=blue,
    urlcolor=blue
}
\begin{document}
\bstctlcite{IEEEexample:BSTcontrol}

\title{RDCompGS: Rate–Distortion Optimization for Compression of 3D Gaussian Splatting with Image Codecs}

\author{Huiping~Ren,
        Shuai~Wan,
        Luqian~Bai,
        Zhang~Chen,
        and~Fuzheng~Yang%
\thanks{Huiping Ren, Shuai Wan and Luqian Bai are with the School of Electronics and
Information, Northwestern Polytechnical University, Xi'an 710129, China
(e-mail: huiping\_ren@mail.nwpu.edu.cn; swan@nwpu.edu.cn; Luqianbai@mail.nwpu.edu.cn).}%
\thanks{Zhang Chen is with the School of Electronics and Information,
Northwestern Polytechnical University, Xi'an 710129, China, and also with the
Department of Computer Science, City University of Hong Kong, Hong Kong SAR,
China (e-mail: chenzhang@mail.nwpu.edu.cn).}%
\thanks{Fuzheng Yang is with the School of Telecommunication Engineering,
Xidian University, Xi'an 710071, China
(e-mail: fzhyang@mail.xidian.edu.cn).}%

}

\maketitle
\begin{abstract}

3D Gaussian Splatting (3DGS) enables high-quality, real-time novel view synthesis but requires significant storage and transmission resources.
Mapping Gaussian attributes to 2D representations enables efficient compression using mature image codecs. 
However, Gaussian representations that achieve comparable rendering quality are generally treated similarly during optimization, despite substantial differences in coding rate, leading to limited rate-distortion (R-D) performance after compression.
To address this limitation, we propose RDCompGS, a 3DGS compression method that enables overall R-D optimization of Gaussian representations with respect to subsequent coding.
A differentiable estimate of the coding rate is incorporated into the R-D optimization with rendering distortion.
The rate constraint adjustment model further regulates the R-D trade-off across different compression conditions.
Extensive experiments on three benchmark datasets demonstrate that RDCompGS achieves average Bjøntegaard Delta Rate reductions of 37.84\% and 49.31\% with H.265/HEVC and JPEG XL, respectively. The estimated rate also shows a strong correlation with the actual bit-rate under different compression conditions.
 
\end{abstract}


\section{Introduction}
\IEEEPARstart{3}{D} Gaussian Splatting (3DGS)~\cite{kerbl20233d} is an important approach for novel view synthesis. By representing scenes with explicit anisotropic 3D Gaussians and rendering them through $\alpha$-blending, 3DGS achieves high rendering quality and real-time performance. 3DGS is an active research topic and is relevant to areas such as autonomous driving, multimedia interaction, virtual reality, cultural heritage digitization, and digital twinning~\cite{tian2026hfrs,shuai2026adversarial,Liang4DGS,zhao2026msa,FieldGS}.
However, representing a 3D scene with 3DGS typically requires millions of Gaussian primitives, leading to significant storage and memory overhead~\cite{ali2026survey,chen2026survey}. Existing studies on 3DGS compression mainly focus on two aspects. 

The first aspect directly compresses the native 3D Gaussian representation. We broadly classify current approaches into Gaussian parameter and attribute compression~\cite{lee2024compact,niedermayr2024compressed,wang2024end,Girish2024EAGLES,chen2026pcgs,chen2025fast,huang2026entropygs,papantonakis2024reducing,fan2024lightgaussian}, anchor-based compression~\cite{lu2024scaffold,chen2024hac,chen2025hacpp,wang2024contextgs,zhan2025cat}, and Gaussian primitive compactification~\cite{hanson2025pup,fan2024lightgaussian,shuai2026adversarial,freitas2026visibility,chen2026refine,Chen2026SPAREGSSP}. These strategies are not mutually exclusive and are often combined to reduce redundancy.

\begin{figure}[!t]
    \centering
    \includegraphics[width=\columnwidth]{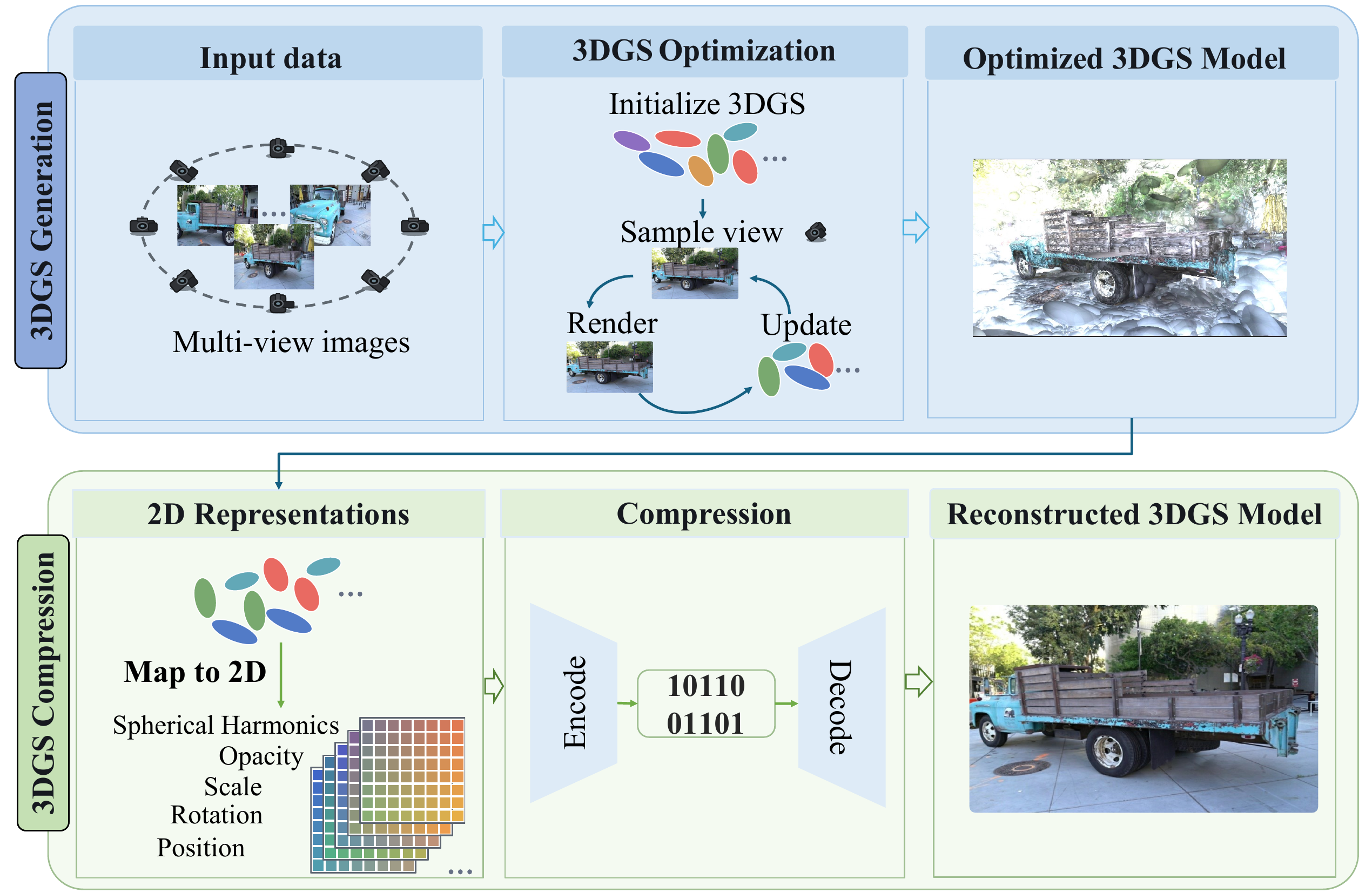}
    \caption{General pipeline for converting 3DGS into 2D representations and compressing them using image codecs. 
    }
    \label{fig:codec_pipeline}
\end{figure}

The second aspect converts 3DGS into 2D representations and compresses them using mature image codecs~\cite{yang2026lightweight,zhu2026Dual,wang20253dgcoding,lee2025compression,morgenstern2024compact,wang2024v}. 
This direction enables 3DGS compression to directly leverage established coding standards and their existing software and hardware support, providing a practical solution for 3DGS storage and transmission. 
As shown in Fig.~\ref{fig:codec_pipeline}, the general pipeline consists of 3DGS generation and compression. The scene is first represented by Gaussian primitives and then mapped into 2D representations for coding.

However, in 3DGS compression through image codecs, Gaussian optimization is usually separated from subsequent coding. Specifically, Gaussian optimization evaluates attributes according to their contribution to rendered views, while
subsequent compression is performed considering both the coding distortion and the corresponding bit-rate as the cost. As the result, Gaussian representations with comparable rendering quality are generally treated similarly during optimization even when their coding rates differ substantially. Therefore, the rate-distortion (R-D) performance cannot be optimized for compression if reducing the rendering distortion is the optimization goal. Since the actual bit-rate produced by the codec is also not differentiable, it is not feasible to be directly incorporated into Gaussian optimization.

Moreover, the balance between coding rate and rendering distortion changes across compression conditions. Therefore, applying a single rate constraint to all compression conditions cannot maintain an appropriate R-D trade-off, which limits the R-D performance after compression.

To address the above issues, we propose RDCompGS, a 3DGS compression method that enables overall R-D optimization in which Gaussian generation and subsequent coding are treated as a whole.
During Gaussian optimization, rendering distortion and the coding rate associated with subsequent compression are jointly optimized to improve the R-D performance after compression.
To incorporate the coding rate into Gaussian optimization, RDCompGS introduces differentiable rate estimation based on the statistical distribution of transformed coefficients and their quantization probabilities.
Different from prior approaches that optimize under a prescribed rate or distortion setting, the rate constraint adjustment model characterizes the relative strength of the estimated rate and rendering distortion and adjusts their balance across different compression conditions.
Extensive experiments on three benchmark datasets demonstrate that RDCompGS achieves average BD-Rate reductions of 37.84\% and 49.31\% with H.265/HEVC and JPEG XL, respectively, while maintaining comparable rendering quality.

The main contributions of this work are summarized as follows.
\begin{itemize}

\item We consider Gaussian generation and subsequent coding as a whole for 3DGS compression and formulate the compression of Gaussian representations as an overall R-D optimization with respect to subsequent
coding.

\item Based on this formulation, we propose RDCompGS, which derives a differentiable estimate of the coding rate from transformed Gaussian attributes and introduces the rate constraint adjustment model to regulate the rate constraint under different coding conditions.

\item We validate RDCompGS on three benchmark datasets using both H.265/HEVC and JPEG XL, showing consistent R-D improvements across different coding conditions. Additional experiments combining Gaussian pruning with image codec compression further confirm the effectiveness of RDCompGS.

\end{itemize}

The rest of this paper is organized as follows.
Section~\ref{sec:related} presents the related work.
Section~\ref{sec:method} introduces the proposed method. Section~\ref{sec:experiments} reports the experimental results and analysis. Finally, Section~\ref{sec:conclusion} concludes this paper.

\section{Related Work}
\label{sec:related}

Based on the Gaussian representation used during compression, we divide existing 3DGS compression methods into two types: compression of 3D Gaussian representations and compression of 3DGS with 2D representations.

\subsection{Compression of 3D Gaussian Representations}
\label{subsec:compact_3d}

We broadly classify methods for compressing 3D Gaussian representations into Gaussian parameter and attribute compression, anchor-based compression, and Gaussian primitive compactification.

Gaussian parameter and attribute compression reduces storage by making the representation of individual Gaussians more compact while retaining the native 3D Gaussian structure.
Representative approaches include quantization, which reduces storage by lowering precision~\cite{lee2024compact,niedermayr2024compressed,wang2024end,Girish2024EAGLES,chen2026pcgs}, entropy coding based on symbol probabilities to reduce the coded length~\cite{chen2025fast,huang2026entropygs}, and spherical harmonics (SH) reduction~\cite{fan2024lightgaussian,papantonakis2024reducing}.

Anchor-based methods reduce 3DGS redundancy by organizing local Gaussians around sparse anchors while exploiting relationships among anchors and their associated attributes. Representative methods include Scaffold-GS~\cite{lu2024scaffold}, HAC~\cite{chen2024hac}, HAC++~\cite{chen2025hacpp}, ContextGS~\cite{wang2024contextgs}, and CAT-3DGS~\cite{zhan2025cat}, which exploit attribute prediction, contextual dependencies, hierarchical relationships, and spatial correlations for compression.

Gaussian primitive compactification reduces redundancy by controlling the number or distribution of Gaussians. Representative methods include PUP 3D-GS~\cite{hanson2025pup}, LightGaussian~\cite{fan2024lightgaussian}, APNet~\cite{shuai2026adversarial}, visibility-based geometry pruning~\cite{freitas2026visibility}, REFINE~\cite{chen2026refine}, and SPARE-GS~\cite{Chen2026SPAREGSSP}. These methods exploit spatial sensitivity scores, global significance scores, view-dependent masks and Rasterization Weight Sampling, multi-view visibility, rendering-aware importance, and global budget allocation to reduce redundant Gaussians or regulate their spatial distribution.

Although these methods effectively reduce redundancy in native 3D Gaussian representations, they mainly compress Gaussian primitives in their original representation form. Another direction incorporates 2D representations into 3DGS compression, enabling the use of established image codecs for compression.

\subsection{Compression of 3DGS with 2D Representations}
\label{subsec:codec_based}

We broadly divide existing 3DGS compression methods based on 2D representations according to when the 2D representation is introduced: during or after the Gaussian optimization process.

Methods that introduce 2D representations after Gaussian optimization adopt different mapping strategies.
LGSCV~\cite{yang2026lightweight} first applies a 3D Morton scan to order Gaussian primitives and then uses a 2D Morton scan to map them into blockwise 2D maps, which are compressed using H.265/HEVC. To improve performance at medium and low bit-rates, it further applies Principal Component Analysis to SH coefficients and introduces MiniPLAS for local primitive permutation. 
Zhu et al.~\cite{zhu2026Dual} propose a Dual-Hilbert scan algorithm that uses a 3D Hilbert curve to preserve spatial locality and a 2D Hilbert curve to map the ordered Gaussian primitives into 2D grids for subsequent H.265/HEVC compression.
3DGCoding~\cite{wang20253dgcoding} sorts Gaussian primitives in z-order according to their positions, maps the quantized attributes onto 2D planes, and compresses the attribute planes of static Gaussians using lossless PNG and those of dynamic Gaussians using H.265/HEVC.
V$^{3}$~\cite{wang2024v} sorts Gaussian primitives using Morton order, maps Gaussian attributes into video frames, and compresses them using H.264/AVC.

For methods that introduce 2D representations during Gaussian optimization, CodecGS~\cite{lee2025compression} redesigns the explicit Gaussian representation using progressive tri-plane features and employs a multilayer perceptron to predict Gaussian attributes. The feature plane channels are then concatenated into frames and compressed using H.265/HEVC. Compared with CodecGS, SOG~\cite{morgenstern2024compact} retains explicit Gaussian attributes and directly organizes them into aligned 2D grids using Parallel Linear Assignment Sorting (PLAS). 

Specifically, PLAS improves the local smoothness of the grids by constructing a smoothed target with a Gaussian filter and performing parallel blockwise assignments to update the arrangement of Gaussian primitives. The optimized Gaussian parameters are then quantized and compressed using JPEG XL.

To summarize, organizing Gaussian parameters into 2D representations provides a structured form that can be compressed using image codecs. 
However, Gaussian optimization remains separated from subsequent coding, which limits the overall R-D performance. 

\section{Methodology}
\label{sec:method}

\begin{figure*}[ht]
    \centering
    \includegraphics[width=\textwidth]{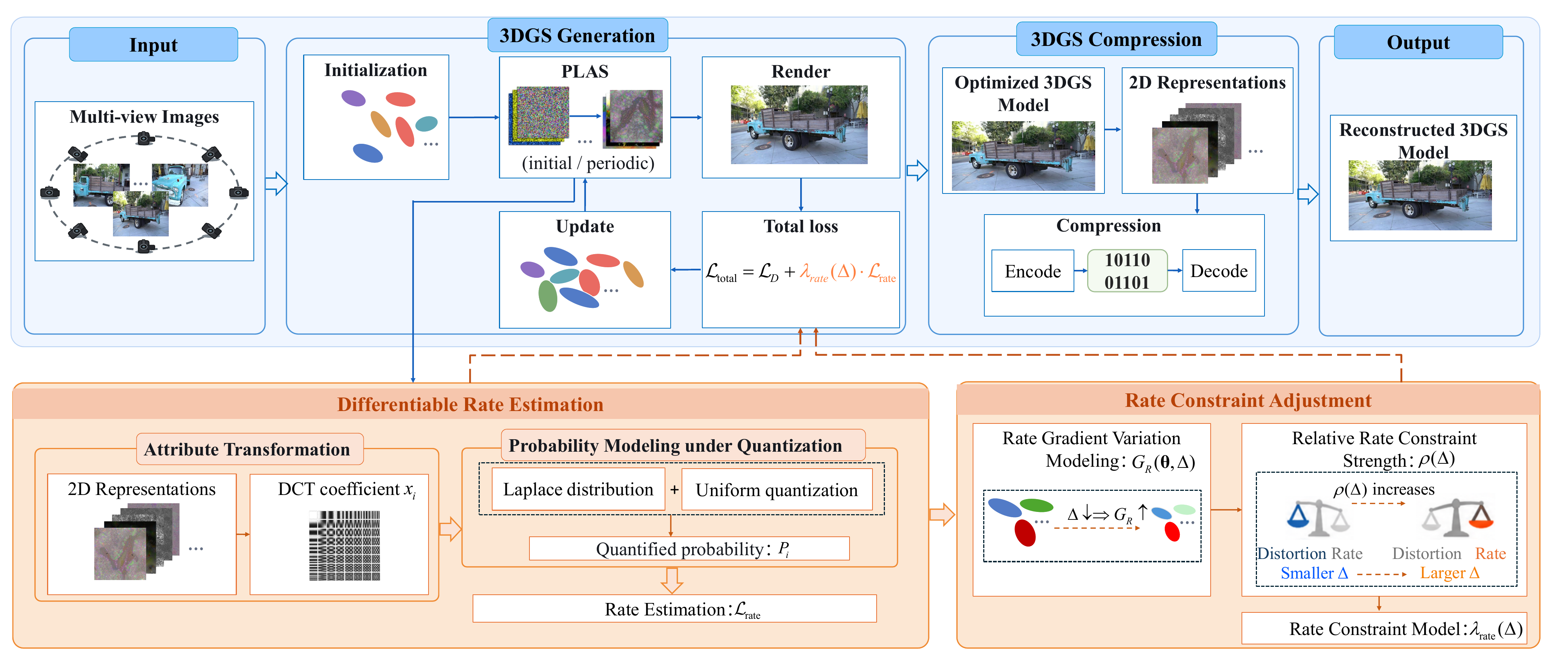}
    \caption{
    \textbf{Overview of the proposed RDCompGS.}
    To enable overall R-D optimization of Gaussian representations with respect to subsequent coding, RDCompGS first introduces \textbf{differentiable rate estimation} based on the distribution of transformed coefficients and their quantization probabilities.
    Then, \textbf{the rate constraint adjustment model} regulates the rate constraint across different compression conditions.
    Finally, the estimated rate and rendering distortion jointly guide Gaussian optimization for subsequent coding.
    }
    \label{fig:framework}
\end{figure*}

To address the mismatch between Gaussian attribute optimization and subsequent coding and the limitation of using a single rate constraint across different coding conditions, we propose RDCompGS, as shown in Fig.~\ref{fig:framework}.
RDCompGS introduces two main components. (a) Differentiable rate estimation constructs a differentiable rate term for Gaussian optimization (Section~\ref{subsec:rate_loss}).
(b) The rate constraint adjustment model regulates the influence of the estimated rate under different coding conditions (Section~\ref{subsec:rate_weight}).
The weighted rate term is then combined with the rendering distortion for Gaussian optimization (Section~\ref{subsec:total_loss}).
The optimized Gaussian attributes are subsequently compressed using image codecs.

\subsection{Differentiable Rate Estimation Model}
\label{subsec:rate_loss}

\begin{figure}[!t]
    \centering
    \includegraphics[width=\columnwidth]{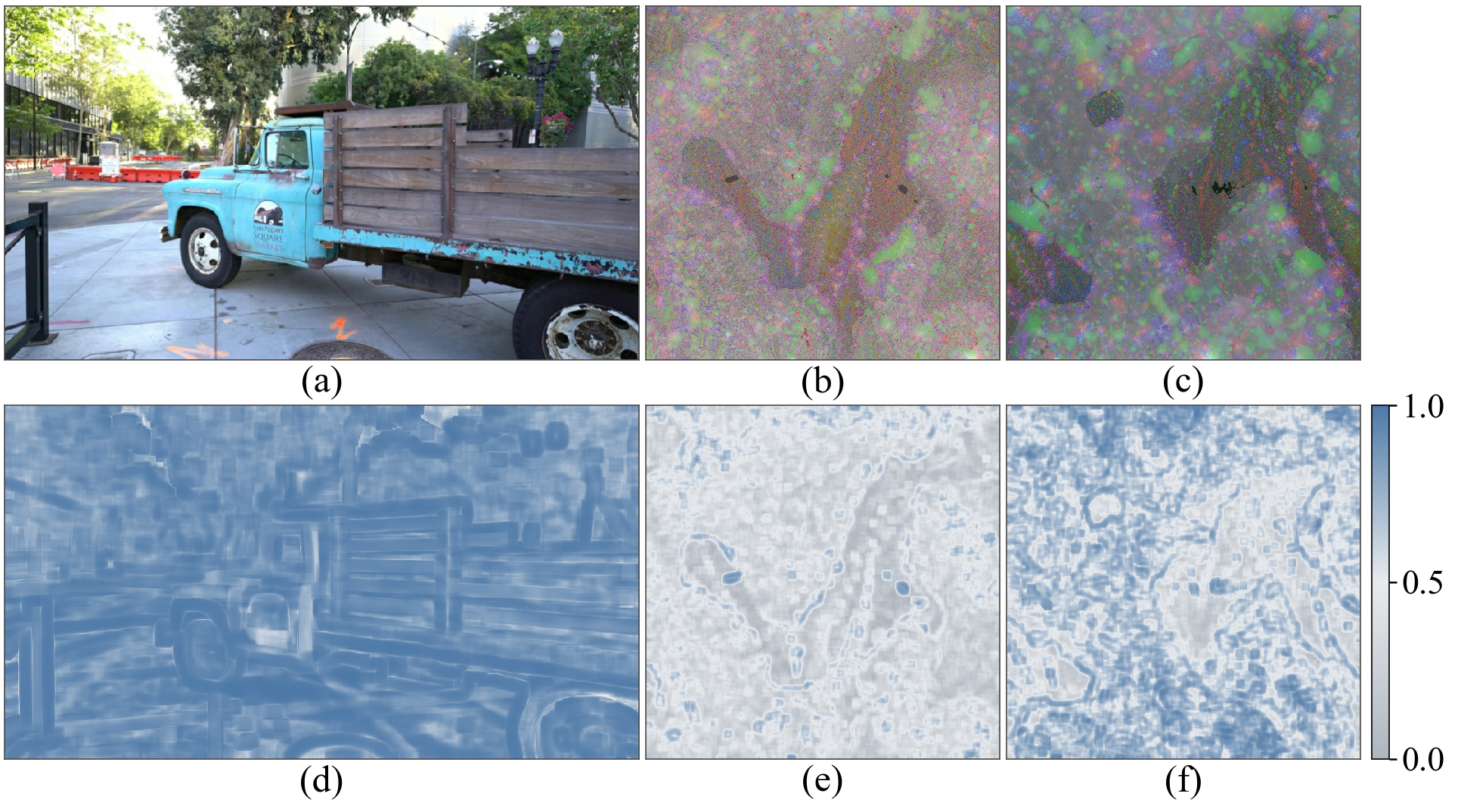}
    \caption{Spatial correlation comparison between a natural image and Gaussian attribute grids on the Truck scene.
    (a) Natural image; (b) and (c) scale attribute grids without and with RDCompGS, respectively; (d), (e), and (f) spatial correlation maps of
    (a), (b), and (c), respectively. Spatial correlation is calculated within
    a $21\times21$ window by averaging the horizontal and vertical Pearson correlations, with gray to blue indicating increasing correlation. The scale attribute shows lower spatial correlation than the natural image, while higher correlation is observed with RDCompGS.}
    \label{fig:attribute_Spatial_correlation}
\end{figure}

Natural images generally exhibit strong spatial correlation, which is exploited by image codecs to reduce spatial redundancy. PLAS organizes Gaussian attributes into 2D grids and places Gaussians with similar attribute values at nearby positions.
As shown in Fig.~\ref{fig:attribute_Spatial_correlation}, PLAS improves the spatial correlation of the Gaussian attribute grids, but a difference from natural images still remains. To further reduce redundancy for subsequent compression while maintaining rendering quality, RDCompGS performs overall R-D optimization of Gaussian representations with respect to subsequent coding.

\begin{figure}[!t]
    \centering
    \includegraphics[width=0.48\linewidth]{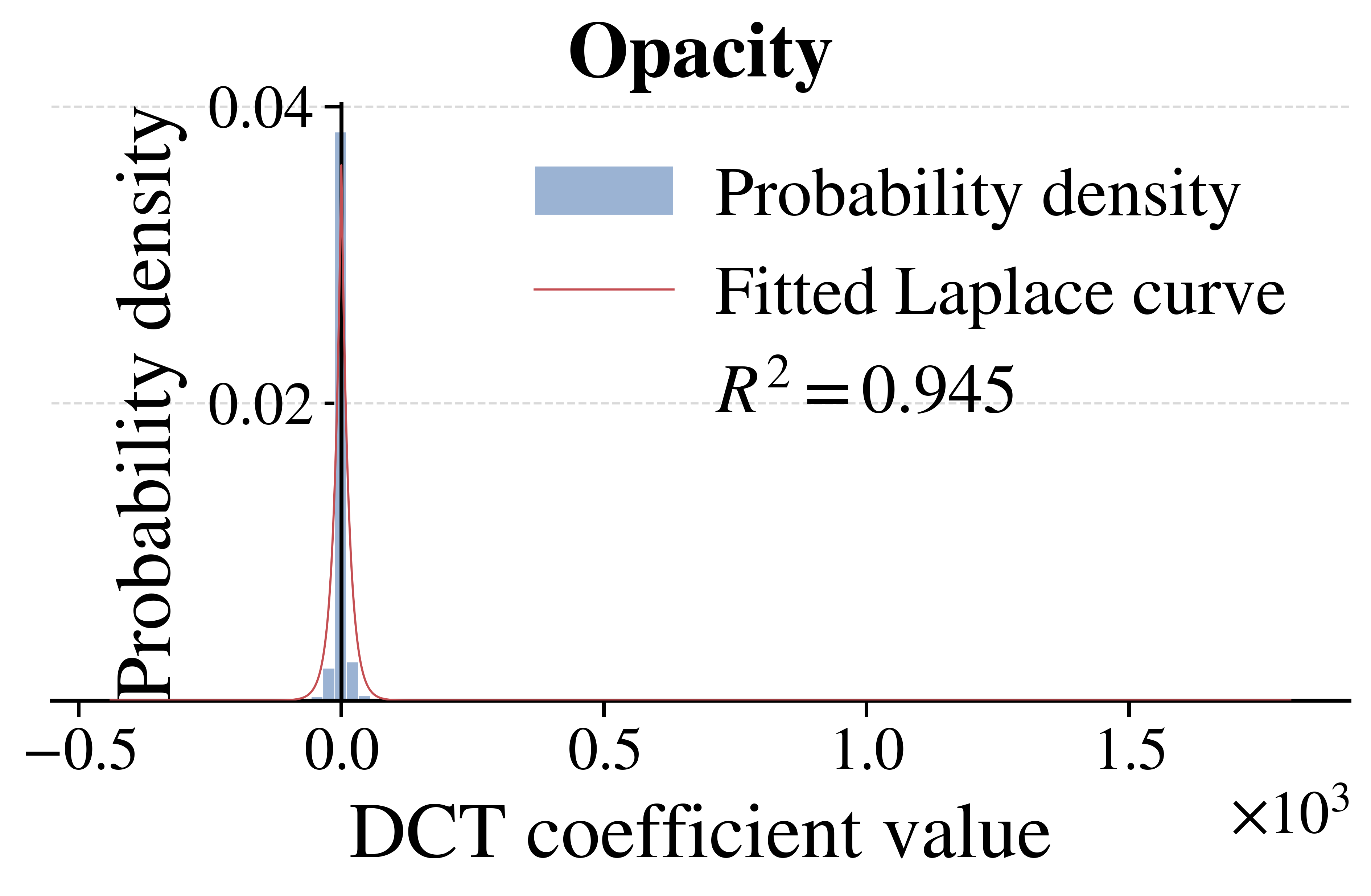}
    \hfill
    \includegraphics[width=0.48\linewidth]{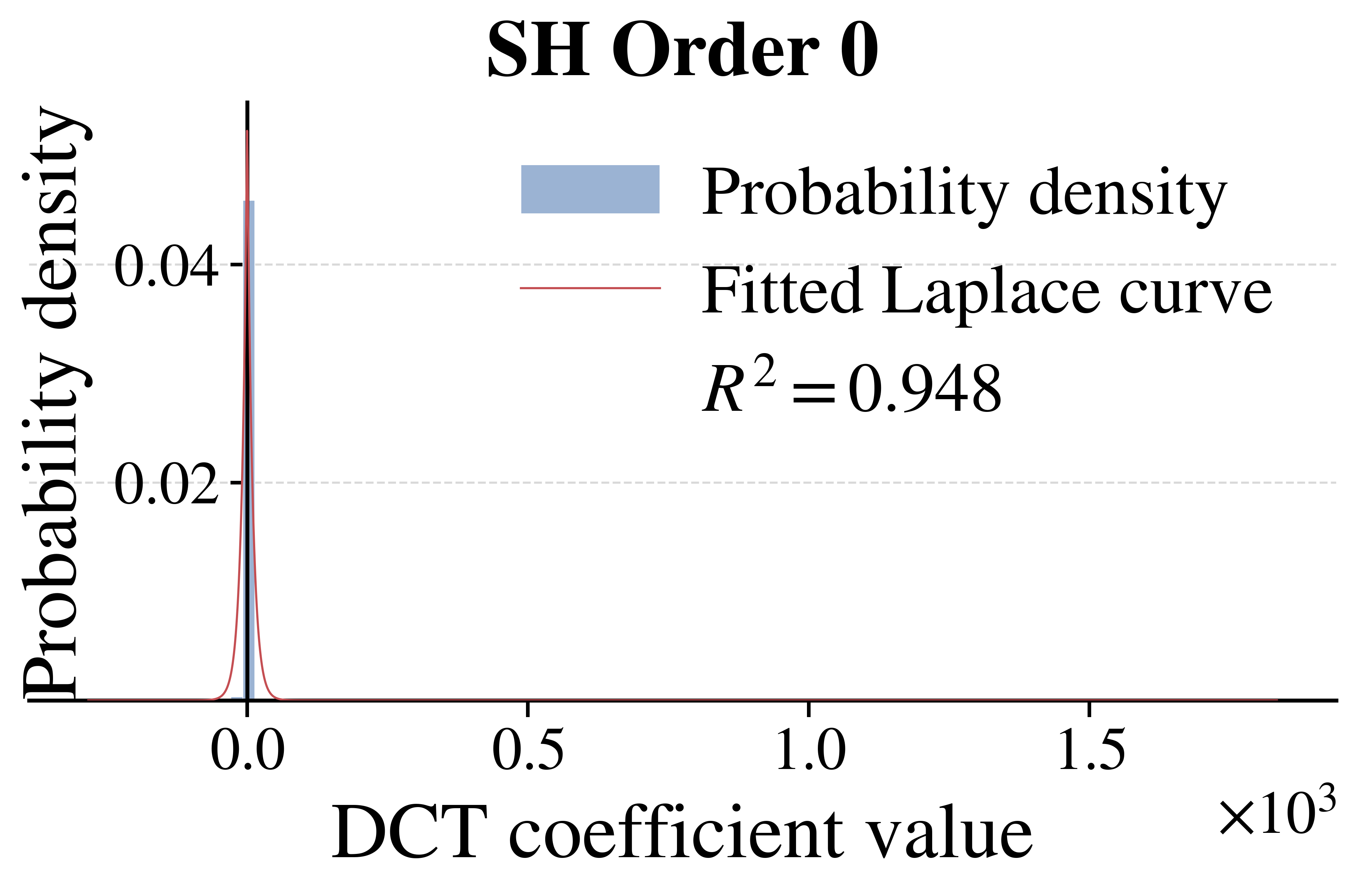}
    \caption{Laplace distribution fitting of DCT coefficients for opacity and SH DC component during training on the Truck scene.}
    \label{fig:DCT}
\end{figure}

To incorporate the coding rate into the overall R-D optimization, we introduce differentiable rate estimation for the non-position Gaussian attributes.
The non-position attributes are first normalized and transformed using an \(8\times8\) block-based Discrete Cosine Transform (DCT), while the position attributes are excluded to preserve scene geometry. 
The statistical analysis in~\cite{lam2000mathematical} motivates modeling the transformed coefficients using a zero-mean Laplace distribution. Fig.~\ref{fig:DCT} further shows that the transformed coefficients exhibit distributions that are consistent with the Laplace model.

Let $x_i$ denote the $i$-th DCT coefficient and $b_i>0$ its scale parameter.
The Laplace probability density function is
\begin{equation}
\label{eq:laplace_pdf}
p(x_i\mid b_i)
=
\frac{1}{2b_i}
\exp\left(
-\frac{|x_i|}{b_i}
\right).
\end{equation}

For a uniform scalar quantizer with step size $\Delta$, the quantized
coefficient is
\begin{equation}
\label{eq:quantized_coefficient}
q_i
=
\operatorname{round}
\left(
\frac{x_i}{\Delta}
\right).
\end{equation}
The probability of obtaining $q_i$ is
\begin{equation}
\label{eq:quantized_probability}
P_i
=
\int_{(q_i-\frac{1}{2})\Delta}^{(q_i+\frac{1}{2})\Delta}
p(t\mid b_i)\,\mathrm{d}t.
\end{equation}

Under the high-resolution scalar quantization approximation, the probability
density is approximately constant within each quantization interval.
Therefore, Eq.~\eqref{eq:quantized_probability} can be approximated as
\begin{equation}
\label{eq:quantized_probability_approx}
P_i
\approx
\Delta p(x_i\mid b_i).
\end{equation}

Based on self-information, we define
$r_i=-\log_2 P_i$.
Using Eqs.~\eqref{eq:quantized_probability_approx} and
\eqref{eq:laplace_pdf}, we obtain
\begin{equation}
\label{eq:rate_expression}
r_i
\approx
\log_2\left(
\frac{2b_i}{\Delta}
\right)
+
\frac{|x_i|}{b_i\ln 2}.
\end{equation}

Since the scale parameter $b_i$ is unknown during Gaussian optimization, we
eliminate it by minimizing Eq.~\eqref{eq:rate_expression} with respect to $b_i$.
For $x_i\neq0$, the minimizing solution is
\begin{equation}
\label{eq:optimal_scale}
b_i^{\star}
=
|x_i|.
\end{equation}
Substituting Eq.~\eqref{eq:optimal_scale} into Eq.~\eqref{eq:rate_expression} yields
\begin{equation}
\label{eq:optimal_rate_expression}
r_i
\approx
\log_2\left(
\frac{2e|x_i|}{\Delta}
\right).
\end{equation}

To make the rate term differentiable and nonnegative during Gaussian optimization, we define
\begin{equation}
\label{eq:coefficient_rate}
r_i
=
\log_2\left(
1+
\frac{2e\sqrt{x_i^2+\epsilon}}{\Delta}
\right),
\qquad
\epsilon=10^{-6}.
\end{equation}

The estimated rate is obtained by accumulating $r_i$ over the DCT coefficients
of the non-position Gaussian attributes:
\begin{equation}
\label{eq:rate_loss}
\mathcal{L}_{\mathrm{rate}}
=
k\sum_{i=1}^{N}r_i
=
k\sum_{i=1}^{N}
\log_2\left(
1+
\frac{2e\sqrt{x_i^2+\epsilon}}{\Delta}
\right),
\end{equation}
where $N$ denotes the total number of DCT coefficients included in the rate
term, and $k$ is a conversion coefficient that scales the accumulated rate to
kbps.

\subsection{The Rate Constraint Adjustment Model}
\label{subsec:rate_weight}

\begin{figure*}[!t]
\centering
\includegraphics[width=0.9 \textwidth]
{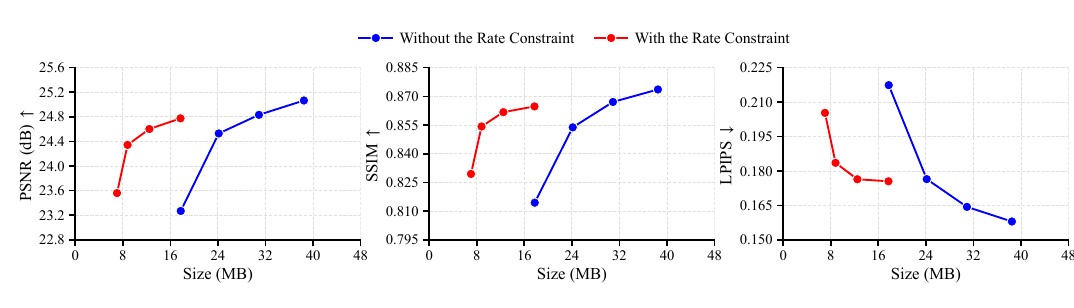}
\caption{R-D performance comparison on the Truck scene under H.265/HEVC compression between Gaussian optimization without the rate constraint and with the rate constraint using $\lambda_{\mathrm{rate}}=1\times10^{-7}$ at QP values of 4, 10, 16, and 22.}
\label{fig:same_weight}
\end{figure*}

During Gaussian optimization, the rendering distortion and the weighted estimated rate are jointly used to update the Gaussian parameters.
Since the gradient magnitude of the estimated rate varies with the quantization step, a single rate constraint cannot maintain an appropriate balance between the estimated rate and rendering distortion across different compression conditions. As illustrated in Fig.~\ref{fig:same_weight}, using a single rate constraint provides only limited improvements in the high-bit-rate region. RDCompGS therefore introduces the rate constraint adjustment model to adjust the rate constraint according to the quantization step.

We first analyze how the quantization step affects the rate gradient that is
backpropagated to the Gaussian parameters. According to
Eq.~\eqref{eq:rate_loss}, the gradient of the estimated rate with respect to the
$i$-th DCT coefficient is
\begin{equation}
\label{eq:rate_coefficient_gradient}
\frac{\partial \mathcal{L}_{\mathrm{rate}}}{\partial x_i}
=
\frac{
2ekx_i
}{
\ln 2\,
\sqrt{x_i^2+\epsilon}
\left(
\Delta+2e\sqrt{x_i^2+\epsilon}
\right)
}.
\end{equation}
For a fixed DCT coefficient $x_i$,
Eq.~\eqref{eq:rate_coefficient_gradient} shows that increasing the quantization
step reduces the gradient magnitude. When
\begin{equation}
2e\sqrt{x_i^2+\epsilon}\ll\Delta,
\end{equation}
the dependence on $\Delta$ can be approximated as
\begin{equation}
\label{eq:coefficient_gradient_scaling}
\left|
\frac{\partial \mathcal{L}_{\mathrm{rate}}}{\partial x_i}
\right|
\propto
\Delta^{-1}.
\end{equation}

Let $\boldsymbol{\theta}$ denote the Gaussian parameters and let $\mathbf{x}=F(\boldsymbol{\theta})$ denote the DCT coefficients obtained from the non-position Gaussian parameters. By the chain rule,
\begin{equation}
\label{eq:rate_gradient_chain}
\nabla_{\boldsymbol{\theta}}
\mathcal{L}_{\mathrm{rate}}
=
\mathbf{J}_{F}^{T}
\nabla_{\mathbf{x}}
\mathcal{L}_{\mathrm{rate}},
\end{equation}
where $\mathbf{J}_{F}$ denotes the Jacobian of
$F(\boldsymbol{\theta})$ with respect to $\boldsymbol{\theta}$.
Since the mapping from the Gaussian parameters to the DCT coefficients does
not explicitly depend on $\Delta$, the dependence of
$\nabla_{\boldsymbol{\theta}}\mathcal{L}_{\mathrm{rate}}$ on the
quantization step is introduced through
$\nabla_{\mathbf{x}}\mathcal{L}_{\mathrm{rate}}$.
According to Eq.~\eqref{eq:coefficient_gradient_scaling}, the rate gradient with
respect to the Gaussian parameters therefore approximately satisfies
\begin{equation}
\label{eq:parameter_gradient_scaling}
\nabla_{\boldsymbol{\theta}}
\mathcal{L}_{\mathrm{rate}}
(\boldsymbol{\theta},\Delta)
\approx
\frac{\Delta_{\mathrm{ref}}}{\Delta}
\nabla_{\boldsymbol{\theta}}
\mathcal{L}_{\mathrm{rate}}
(\boldsymbol{\theta},\Delta_{\mathrm{ref}}),
\end{equation}
where $\Delta_{\mathrm{ref}}$ denotes a reference quantization step.

To characterize the magnitude of the rate gradient, we define
\begin{equation}
\label{eq:rate_gradient_norm}
G_R(\boldsymbol{\theta},\Delta)
=
\left\lVert
\nabla_{\boldsymbol{\theta}}
\mathcal{L}_{\mathrm{rate}}
(\boldsymbol{\theta},\Delta)
\right\rVert_2.
\end{equation}
The corresponding magnitude at $\Delta_{\mathrm{ref}}$ is denoted by
\begin{equation}
\label{eq:reference_rate_gradient}
G_{R,\mathrm{ref}}
=
G_R(\boldsymbol{\theta},\Delta_{\mathrm{ref}}).
\end{equation}
Taking the $L_2$ norm on both sides of
Eq.~\eqref{eq:parameter_gradient_scaling} gives
\begin{equation}
\label{eq:rate_gradient_scaling}
\frac{
G_R(\boldsymbol{\theta},\Delta)
}{
G_{R,\mathrm{ref}}
}
\approx
\frac{
\Delta_{\mathrm{ref}}
}{
\Delta
}.
\end{equation}

The same Gaussian parameters are also updated by the rendering distortion. We
therefore define
\begin{equation}
\label{eq:rendering_gradient_norm}
G_D(\boldsymbol{\theta})
=
\left\lVert
\nabla_{\boldsymbol{\theta}}
\mathcal{L}_{D}(\boldsymbol{\theta})
\right\rVert_2,
\end{equation}
where $G_D$ denotes the magnitude of the rendering distortion gradient with respect to the Gaussian parameters.

To characterize the influence of the estimated rate relative to the rendering distortion, we define
\begin{equation}
\label{eq:relative_rate_strength}
\rho(\Delta)
=
\frac{
\lambda_{\mathrm{rate}}(\Delta)
G_R(\boldsymbol{\theta},\Delta)
}{
G_D(\boldsymbol{\theta})
},
\end{equation}
where $\lambda_{\mathrm{rate}}(\Delta)$ denotes the rate constraint at $\Delta$, and $\rho(\Delta)$ represents the relative rate constraint strength with respect to the rendering distortion.
The reference value of the relative rate constraint strength is defined as
\begin{equation}
\label{eq:reference_rate_strength}
\rho_{\mathrm{ref}}
=
\frac{
\lambda_{\mathrm{ref}}
G_{R,\mathrm{ref}}
}{
G_D(\boldsymbol{\theta})
},
\end{equation}
where $\lambda_{\mathrm{ref}}$ denotes the rate constraint at
$\Delta_{\mathrm{ref}}$.

Combining Eqs.~\eqref{eq:relative_rate_strength},
\eqref{eq:reference_rate_strength}, and
\eqref{eq:rate_gradient_scaling} gives
\begin{equation}
\label{eq:relative_rate_ratio}
\frac{
\rho(\Delta)
}{
\rho_{\mathrm{ref}}
}
\approx
\frac{
\lambda_{\mathrm{rate}}(\Delta)
}{
\lambda_{\mathrm{ref}}
}
\frac{
\Delta_{\mathrm{ref}}
}{
\Delta
}.
\end{equation}

If the rate constraint is adjusted only to compensate for the reduction in the rate gradient magnitude as $\Delta$ increases, $\rho(\Delta)$ should remain approximately unchanged. However, this does not reflect the different compression requirements across quantization steps. A smaller $\Delta$ places more emphasis on rendering quality, whereas a larger $\Delta$ requires greater emphasis on reducing redundancy in Gaussian attributes. Accordingly, we let $\rho(\Delta)$ increase with $\Delta$ as

\begin{equation}
\label{eq:target_rate_strength}
\frac{
\rho(\Delta)
}{
\rho_{\mathrm{ref}}
}
=
\left(
\frac{
\Delta
}{
\Delta_{\mathrm{ref}}
}
\right)^{\beta},
\qquad
\beta > 0,
\end{equation}
where $\beta$ controls the sensitivity of $\rho(\Delta)$ to the quantization step.

Combining Eq.~\eqref{eq:target_rate_strength} with
Eq.~\eqref{eq:relative_rate_ratio}, the rate constraint is defined as
\begin{equation}
\label{eq:adaptive_rate_weight}
\lambda_{\mathrm{rate}}(\Delta)
=
\lambda_{\mathrm{ref}}
\left(
\frac{\Delta}{\Delta_{\mathrm{ref}}}
\right)^{1+\beta}.
\end{equation}

\begin{algorithm}[ht]
\caption{R-D Optimization in RDCompGS}
\label{alg:rate_guided_training}

\renewcommand{\algorithmicrequire}{\textbf{Input:}}
\renewcommand{\algorithmicensure}{\textbf{Output:}}

\begin{algorithmic}[1]

\REQUIRE Multi-view images, camera parameters, 
quantization step $\Delta$.

\ENSURE Optimized Gaussian parameters $\boldsymbol{\theta}$.

\STATE Initialize the Gaussian parameters $\boldsymbol{\theta}$
using multi-view images and camera parameters.

\STATE Apply PLAS to organize the Gaussian attributes into 2D grids.

\FOR{each training iteration}

\STATE Obtain the rendered image through differentiable
Gaussian rasterization.

\STATE Compute the rendering distortion $\mathcal{L}_{D}$ using
Eq.~\eqref{eq:rendering_loss}.

\IF{the densification stage has ended}

\STATE \textit{Part 1: Differentiable Rate Estimation
(Section~\ref{subsec:rate_loss})}

\STATE Exclude the position attributes and normalize the remaining
2D representations.

\STATE Apply $8\times8$ block-based DCT to obtain the coefficients
$\{x_i\}_{i=1}^{N}$.

\STATE Compute the estimated rate
$\mathcal{L}_{\mathrm{rate}}$ using
Eq.~\eqref{eq:rate_loss}.

\STATE \textit{Part 2: The Rate Constraint Adjustment Model
(Section~\ref{subsec:rate_weight})}

\STATE Compute the rate constraint
$\lambda_{\mathrm{rate}}(\Delta)$ using
Eq.~\eqref{eq:adaptive_rate_weight}.

\STATE \textit{Part 3: Total Training Loss
(Section~\ref{subsec:total_loss})}

\STATE Compute the total training loss
$\mathcal{L}_{\mathrm{total}}$ using
Eq.~\eqref{eq:total_loss}.

\ELSE

\STATE Set
$\mathcal{L}_{\mathrm{total}}
\leftarrow
\mathcal{L}_{D}$.

\ENDIF

\STATE Backpropagate $\mathcal{L}_{\mathrm{total}}$ and update
$\boldsymbol{\theta}$.

\IF{a densification step is performed}

\STATE Reapply PLAS to restore the local structure of the
2D representations.

\ENDIF

\ENDFOR

\RETURN Optimized Gaussian parameters $\boldsymbol{\theta}$.

\end{algorithmic}
\end{algorithm}

\subsection{Total Training Loss}
\label{subsec:total_loss}

Following 3DGS~\cite{kerbl20233d}, the rendering distortion is defined as
\begin{equation}
\label{eq:rendering_loss}
\mathcal{L}_{D}
=
(1-\lambda_{\mathrm{dssim}})\mathcal{L}_{L1}
+
\lambda_{\mathrm{dssim}}\mathcal{L}_{\mathrm{D\text{-}SSIM}},
\end{equation}
where $\lambda_{\mathrm{dssim}}=0.2$.

The distortion considered in RDCompGS refers to the rendering distortion defined above, while the distortion introduced by compression is not explicitly modeled in the optimization process.

Before densification ends, the Gaussian representation is optimized using
only $\mathcal{L}_{D}$. After densification, the estimated rate in
Eq.~\eqref{eq:rate_loss} is incorporated into Gaussian optimization using $\lambda_{\mathrm{rate}}(\Delta)$ in
Eq.~\eqref{eq:adaptive_rate_weight}. The total training loss is
\begin{equation}
\label{eq:total_loss}
\mathcal{L}_{\mathrm{total}}
=
\mathcal{L}_{D}
+
\lambda_{\mathrm{rate}}(\Delta)
\mathcal{L}_{\mathrm{rate}}.
\end{equation}

By jointly incorporating the estimated rate and rendering distortion, RDCompGS connects Gaussian optimization with subsequent coding to achieve overall R-D optimization for 3DGS compression. 
The R-D optimization procedure in RDCompGS is summarized in Algorithm~\ref{alg:rate_guided_training}.

\section{Experiments}
\label{sec:experiments}

\subsection{Experimental Setup}
\label{subsec:experimental_setup}

\subsubsection{Datasets and Evaluation Metrics} 
We conduct experiments on three datasets: Tanks and Temples~\cite{knapitsch2017tanks}, Deep Blending~\cite{hedman2018deep}, and Mip-NeRF 360~\cite{barron2022mip}. These datasets contain diverse indoor and outdoor scenes with complex geometry and texture details. We evaluate rendering quality using Peak Signal-to-Noise Ratio (PSNR), Structural Similarity Index Measure (SSIM)~\cite{wang2004SSIM}, and Learned Perceptual Image Patch Similarity (LPIPS)~\cite{zhang2018LPIPS}. The encoded data sizes generated by H.265/HEVC and JPEG XL are used for compression evaluation. The coding performance is evaluated in terms of the Bjøntegaard delta rate (BD-Rate)~\cite{bjontegaard2001calculation}.

\subsubsection{Implementation Details}
The proposed method is implemented in PyTorch. Experiments are conducted on an NVIDIA GeForce RTX 5070 Ti GPU. During training, PLAS is applied after Gaussian initialization and after each densification step to organize the Gaussian parameters into 2D representations. After the densification stage, differentiable rate estimation is introduced into Gaussian optimization.

Following the settings in V$^{3}$~\cite{wang2024v} and SOG~\cite{morgenstern2024compact}, Gaussian attributes are first quantized and then compressed using image codecs.
The position attributes are quantized to 16-bit values and split into high and low 8-bit components, which are represented as two three-channel RGB images.
The remaining attributes are quantized to 8-bit values.
Specifically, the opacity attribute and the $w$ component of the rotation quaternion are represented as single-channel grayscale images, whereas the scale attribute, the $xyz$ components of the rotation quaternion, and the SH coefficients are represented as three-channel RGB images.
These attribute images are then compressed using H.265/HEVC~\cite{sullivan2012overview} (all intra coding) or JPEG XL~\cite{alakuijala2019jpeg}, where the position attributes are encoded losslessly and the remaining attributes are encoded using lossy compression.

For H.265/HEVC compression, the quantization parameter (QP) values are set to 4, 10, 16, and 22. The corresponding quantization step size $\Delta$ is calculated following~\cite{BudagaviHEVC}.
Higher QP values are not considered because they cause severe degradation in rendering quality. To determine the JPEG XL distance value, we select 260 3DGS attribute images from the three datasets. The position attribute images are excluded. For each H.265/HEVC QP, we use binary search to determine the JPEG XL distance value that produces the same bits per pixel. The resulting distance values are 0.1005, 0.2157, 0.4511, and 0.9363 for QP values of 4, 10, 16, and 22, respectively. 
For SOG, the SH attributes are compressed using JPEG XL with quality levels of 95, 90, 70, and 50, while the default compression settings are used for all other attributes.
The coefficient \(k=1/40\) is derived from the encoder frame rate of 25 frames per second.

Based on the parameter analysis conducted on the Truck scene from Tanks and Temples, we set
$\Delta_{\mathrm{ref}}=8$,
$\lambda_{\mathrm{ref}}=1\times10^{-7}$,
and $\beta=0.2$. Since the Truck scene is used to determine the reference parameters, it is excluded from all subsequent R-D evaluations to avoid potential bias. 
The exceptions are the validation of the rate estimation model in
Section~\ref{subsec:rate_correlation} and the analysis of the rate constraint adjustment model in Section~\ref{subsec:rate_weight_analysis}, where the Truck scene is retained.

\subsection{Compression Performance}
\label{subsec:sog_rd}

We evaluate RDCompGS by compressing the optimized Gaussian representations using JPEG XL and H.265/HEVC, and compare the resulting R-D performance with SOG. Tables~\ref{tab:sog_rdcompgs_jpegxl} and~\ref{tab:sog_rdcompgs_hevc} report the results with JPEG XL and H.265/HEVC, respectively. The BD-Rate results show consistent R-D improvements across the three datasets. With JPEG XL, RDCompGS obtains BD-Rate reductions of 30.44\%, 60.94\%, and 56.55\% on Tanks \& Temples, Deep Blending, and Mip-NeRF 360, respectively, corresponding to an average reduction of 49.31\%. Similar improvements are observed with H.265/HEVC, where the corresponding BD-Rate reductions are 15.44\%, 50.48\%, and 47.61\%, with an average reduction of 37.84\%.

\begin{table*}[!t]
\centering
\caption{R-D performance of RDCompGS using JPEG XL compared with SOG. Size is measured in MB and PSNR in dB.}
\label{tab:sog_rdcompgs_jpegxl}
\footnotesize
\setlength{\tabcolsep}{1.8pt}
\renewcommand{\arraystretch}{1.15}

\begin{tabular}{@{}l c
cccc @{\hspace{4pt}}
cccc @{\hspace{4pt}}
cccc @{\hspace{4pt}}
cccc @{\hspace{4pt}}
c@{}}
\toprule

\multirow{2}{*}{Dataset}
& \multirow{2}{*}{Method}
& \multicolumn{4}{c}{Distance = 0.1005}
& \multicolumn{4}{c}{Distance = 0.2157}
& \multicolumn{4}{c}{Distance = 0.4511}
& \multicolumn{4}{c}{Distance = 0.9363}
& \multirow{2}{*}{\begin{tabular}[c]{@{}c@{}}
BD-Rate\\
(\%) $\downarrow$
\end{tabular}}
\\

\cmidrule(lr){3-6}
\cmidrule(lr){7-10}
\cmidrule(lr){11-14}
\cmidrule(lr){15-18}

&
& Size & PSNR & SSIM & LPIPS
& Size & PSNR & SSIM & LPIPS
& Size & PSNR & SSIM & LPIPS
& Size & PSNR & SSIM & LPIPS
&
\\

\midrule

\multirow{2}{*}{\begin{tabular}[c]{@{}l@{}}
Tanks \& Temples\\[-1pt]
(Train)
\end{tabular}}
& SOG
& 14.828 & 21.54 & 0.794 & 0.231
& 11.343 & 21.36 & 0.791 & 0.238
& 7.779  & 20.78 & 0.778 & 0.250
& 6.896  & 20.25 & 0.766 & 0.257
& \multirow{2}{*}{-30.44}
\\

& RDCompGS
& 16.194 & 21.73 & 0.796 & 0.235
& 10.488 & 21.50 & 0.791 & 0.241
& 6.321  & 21.06 & 0.779 & 0.252
& 4.079  & 20.70 & 0.758 & 0.274
&
\\

\midrule

\multirow{2}{*}{Deep Blending}
& SOG
& 12.175 & 29.22 & 0.891 & 0.276
& 8.969  & 28.93 & 0.889 & 0.280
& 6.338  & 28.15 & 0.882 & 0.288
& 5.845  & 27.68 & 0.879 & 0.292
& \multirow{2}{*}{-60.94}
\\

& RDCompGS
& 11.623 & 29.52 & 0.898 & 0.271
& 7.185  & 29.32 & 0.895 & 0.273
& 4.562  & 29.20 & 0.889 & 0.282
& 3.136  & 28.63 & 0.879 & 0.299
&
\\

\midrule

\multirow{2}{*}{Mip-NeRF 360}
& SOG
& 37.140 & 26.82 & 0.797 & 0.235
& 28.136 & 26.53 & 0.793 & 0.244
& 19.073 & 25.62 & 0.776 & 0.261
& 16.933 & 24.88 & 0.762 & 0.271
& \multirow{2}{*}{-56.55}
\\

& RDCompGS
& 27.284 & 26.91 & 0.799 & 0.243
& 16.545 & 26.70 & 0.792 & 0.250
& 10.608 & 26.30 & 0.782 & 0.261
& 7.459  & 25.45 & 0.755 & 0.285
&
\\

\bottomrule
\end{tabular}
\end{table*}

\begin{table*}[!t]
\centering
\caption{R-D performance of RDCompGS using H.265/HEVC compared with SOG. Size is measured in MB and PSNR in dB.}
\label{tab:sog_rdcompgs_hevc}
\footnotesize
\setlength{\tabcolsep}{2.0pt}
\renewcommand{\arraystretch}{1.15}

\begin{tabular}{@{}l c
cccc @{\hspace{4pt}}
cccc @{\hspace{4pt}}
cccc @{\hspace{4pt}}
cccc @{\hspace{4pt}}
c@{}}
\toprule

\multirow{2}{*}{Dataset}
& \multirow{2}{*}{Method}
& \multicolumn{4}{c}{QP = 4}
& \multicolumn{4}{c}{QP = 10}
& \multicolumn{4}{c}{QP = 16}
& \multicolumn{4}{c}{QP = 22}
& \multirow{2}{*}{\begin{tabular}[c]{@{}c@{}}
BD-Rate\\
(\%) $\downarrow$
\end{tabular}}
\\

\cmidrule(lr){3-6}
\cmidrule(lr){7-10}
\cmidrule(lr){11-14}
\cmidrule(lr){15-18}

&
& Size & PSNR & SSIM & LPIPS
& Size & PSNR & SSIM & LPIPS
& Size & PSNR & SSIM & LPIPS
& Size & PSNR & SSIM & LPIPS
&
\\

\midrule

\multirow{2}{*}{\begin{tabular}[c]{@{}l@{}}
Tanks \& Temples\\[-1pt]
(Train)
\end{tabular}}
& SOG
& 14.828 & 21.54 & 0.794 & 0.231
& 11.343 & 21.36 & 0.791 & 0.238
& 7.779  & 20.78 & 0.778 & 0.250
& 6.896  & 20.25 & 0.766 & 0.257
& \multirow{2}{*}{-15.44}
\\

& RDCompGS
& 16.451 & 21.57 & 0.795 & 0.237
& 11.180 & 21.24 & 0.785 & 0.244
& 6.886  & 21.04 & 0.776 & 0.255
& 4.645  & 20.51 & 0.753 & 0.279
&
\\

\midrule

\multirow{2}{*}{Deep Blending}
& SOG
& 12.175 & 29.22 & 0.891 & 0.276
& 8.969  & 28.93 & 0.889 & 0.280
& 6.338  & 28.15 & 0.882 & 0.288
& 5.845  & 27.68 & 0.879 & 0.292
& \multirow{2}{*}{-50.48}
\\

& RDCompGS
& 11.554 & 29.46 & 0.897 & 0.271
& 7.603  & 29.25 & 0.895 & 0.275
& 4.846  & 29.04 & 0.888 & 0.283
& 3.417  & 28.36 & 0.876 & 0.307
&
\\

\midrule

\multirow{2}{*}{Mip-NeRF 360}
& SOG
& 37.140 & 26.82 & 0.797 & 0.235
& 28.136 & 26.53 & 0.793 & 0.244
& 19.073 & 25.62 & 0.776 & 0.261
& 16.933 & 24.88 & 0.762 & 0.271
& \multirow{2}{*}{-47.61}
\\

& RDCompGS
& 27.351 & 26.90 & 0.798 & 0.244
& 17.282 & 26.61 & 0.791 & 0.251
& 11.322 & 26.02 & 0.777 & 0.264
& 8.204  & 25.12 & 0.748 & 0.291
&
\\

\bottomrule
\end{tabular}
\end{table*}

\begin{figure*}[!t]
    \centering
    \includegraphics[width=0.7 \textwidth]{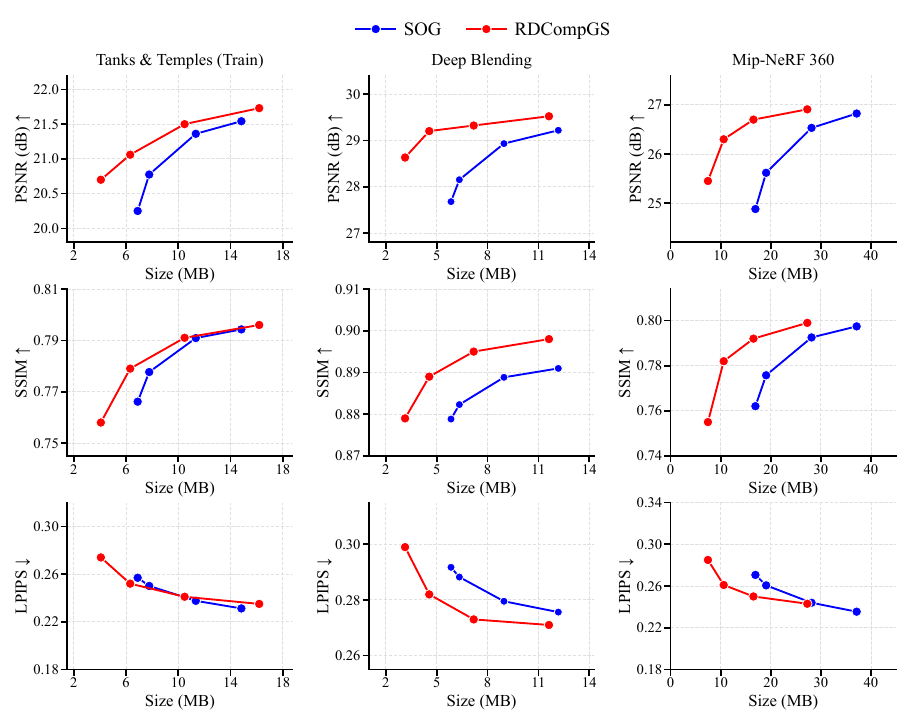}
    \caption{R-D performance comparison between SOG and RDCompGS compressed using JPEG XL on three benchmark datasets in terms of PSNR, SSIM, and LPIPS.}
    \label{fig:sog_rdcompgs_jpegxl}
\end{figure*}

Fig.~\ref{fig:sog_rdcompgs_jpegxl} further compares the R-D performance of SOG and RDCompGS compressed using JPEG XL in terms of PSNR, SSIM, and LPIPS. For PSNR and SSIM,
the RDCompGS curves are generally located above and to the left of those of
SOG, indicating improved R-D performance across the three datasets, although
SOG achieves slightly lower LPIPS in the high-bit-rate region on Tanks \& Temples
and Mip-NeRF 360.

\begin{figure*}[!t]
    \centering
    \includegraphics[width=\textwidth]{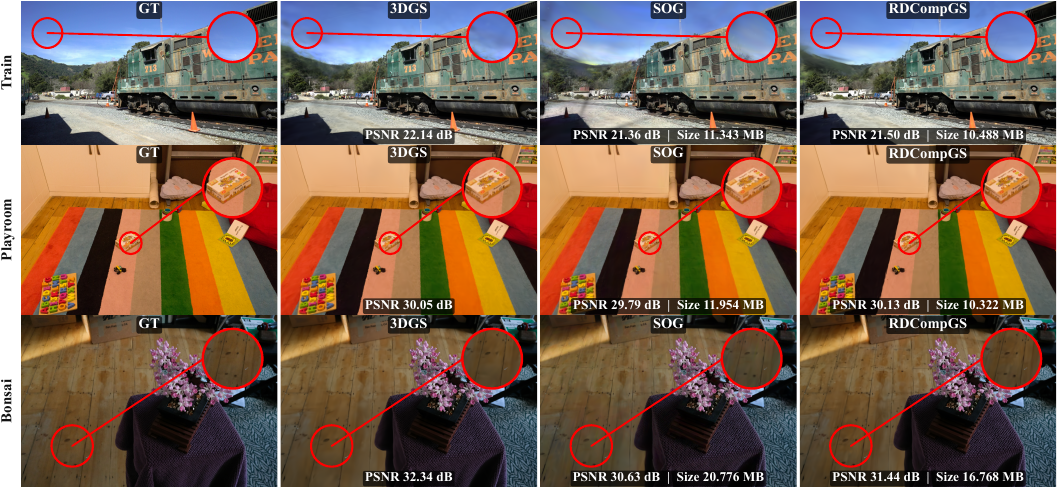}
    \caption{Visual comparison on the Train, Playroom, and Bonsai scenes, where RDCompGS is compressed using JPEG XL. Each row shows GT, 3DGS, SOG, and RDCompGS from left to right, with the same local region enlarged for comparison.}
    \label{fig:qualitative_SOG}
\end{figure*}

Fig.~\ref{fig:qualitative_SOG} presents visual comparisons on the Train, Playroom, and Bonsai scenes. RDCompGS achieves higher PSNR than SOG with smaller compressed
sizes across the three scenes. The enlarged regions further show that RDCompGS better preserves the sky appearance in Train, the paper box details in Playroom, and the floor textures in Bonsai.

\subsection{Validation of Rate Estimation Model}
\label{subsec:rate_correlation}
Having established the R-D improvements, we further evaluate the correlation between the estimated rate and the actual bit-rate. We use 260 non-position Gaussian attribute images generated after PLAS sorting. For each image, the estimated rate is compared with the actual bit-rate obtained by encoding the same image with H.265/HEVC.

\begin{figure}[!t]
    \centering
    \includegraphics[width=\columnwidth]{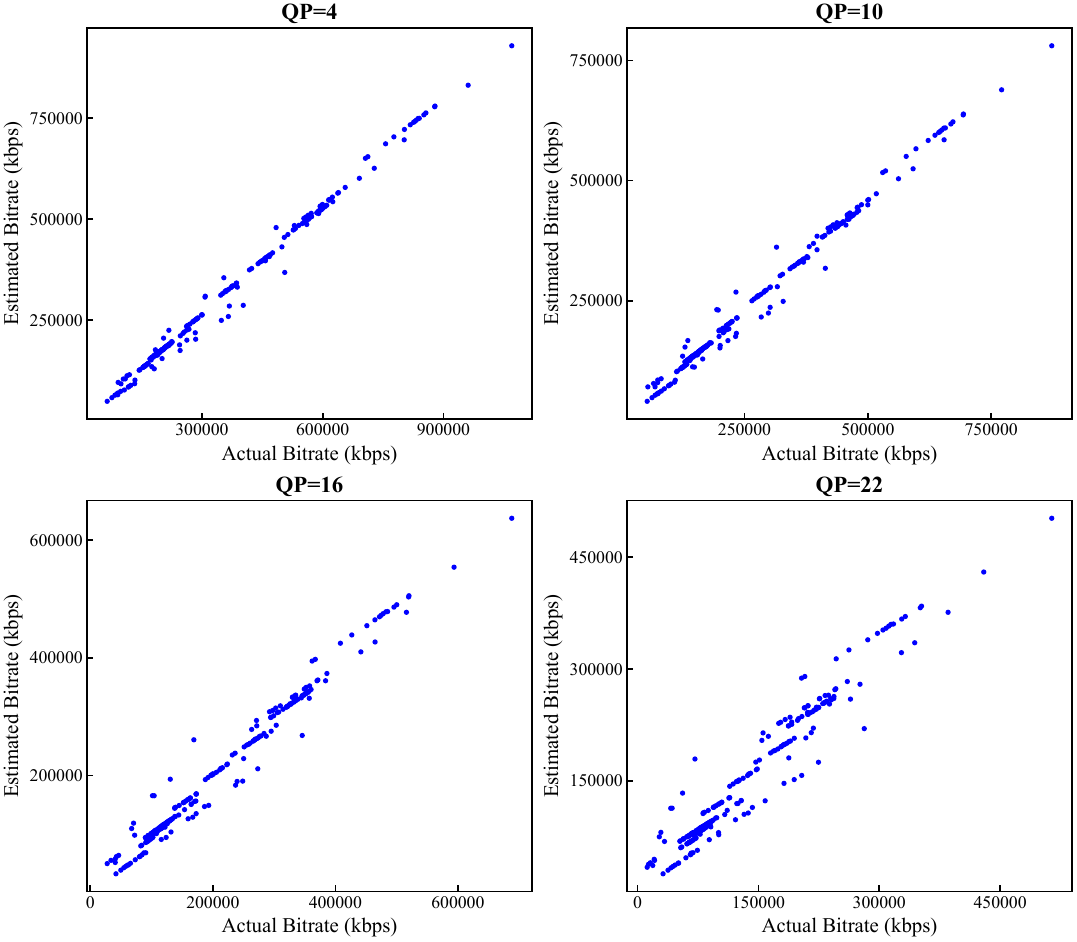}
    \caption{Estimated rate and actual bit-rate produced by H.265/HEVC at different QP values. Each point corresponds to one Gaussian attribute image.}
    \label{fig:Rate_Loss_Estimation_Correlation}
\end{figure}

As shown in Fig.~\ref{fig:Rate_Loss_Estimation_Correlation}, the estimated rate exhibits a strong linear correlation with the actual bit-rate across all tested QP values. At QP values of 4, 10, 16, and 22, the Pearson correlation coefficients are 0.997, 0.996, 0.990, and 0.973, respectively.

\subsection{Analysis of the Rate Constraint Adjustment Model}
\label{subsec:rate_weight_analysis}

The parameter settings of the rate constraint adjustment model are analyzed in this section. The reference quantization step is set to $\Delta_{\mathrm{ref}}=8$. According to Eq.~\eqref{eq:adaptive_rate_weight}, using a relatively large $\Delta_{\mathrm{ref}}$ keeps the rate constraint small at small quantization steps, avoiding an excessive rate constraint in the high-bit-rate region.
The effects of $\lambda_{\mathrm{ref}}$ and $\beta$ are evaluated on the Truck scene, and the selected values are used in the subsequent experiments.

\subsubsection{Selection of
\texorpdfstring{$\lambda_{\mathrm{ref}}$}{lambda ref}}

\begin{figure}[!t]
    \centering
    \includegraphics[width=\columnwidth]{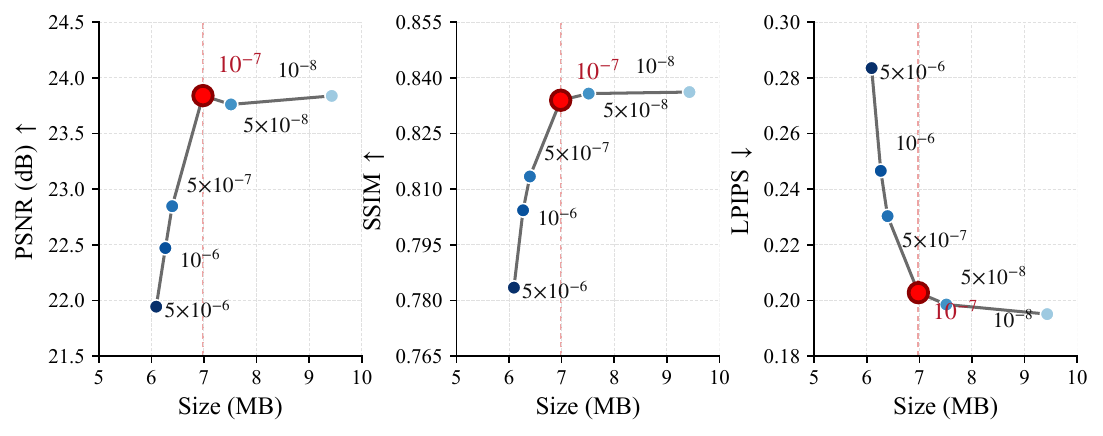}
    \caption{R-D curves on the Truck scene for different $\lambda_{\mathrm{ref}}$ values. The selected value $\lambda_{\mathrm{ref}}=1\times10^{-7}$ achieves a favorable R-D trade-off according to the observed trend of the curves.}
    \label{fig:truck_lambda_ref}
\end{figure}

Fig.~\ref{fig:truck_lambda_ref} presents the R-D curves obtained with
different $\lambda_{\mathrm{ref}}$ values. As
$\lambda_{\mathrm{ref}}$ increases, the compressed size generally decreases,
showing that a larger rate coefficient places a stronger constraint on the
estimated rate during Gaussian optimization. For relatively small
$\lambda_{\mathrm{ref}}$ values, the reduction in compressed size is limited,
while the rendering quality remains relatively stable. Increasing
$\lambda_{\mathrm{ref}}$ to $1\times10^{-7}$ provides a more evident size
reduction without noticeable degradation in the three quality metrics.
Further increasing $\lambda_{\mathrm{ref}}$ results in only limited additional
size reduction, while PSNR and SSIM decrease and LPIPS increases more
noticeably. Considering these trends,
$\lambda_{\mathrm{ref}}=1\times10^{-7}$ provides a favorable trade-off
between compression efficiency and rendering quality and is therefore adopted
in the subsequent experiments.

\subsubsection{Selection of
\texorpdfstring{$\beta$}{beta}}

\begin{figure}[!t]
    \centering
    \includegraphics[width=0.6\columnwidth]{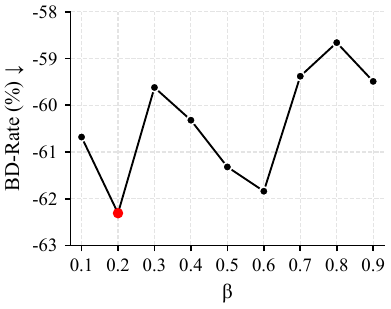}
    \caption{BD-Rate comparison of different $\beta$ values on the Truck scene under H.265/HEVC compression.}
    \label{fig:beta}
\end{figure}

Fig.~\ref{fig:beta} evaluates the sensitivity of the rate constraint adjustment model to different $\beta$ values, using Gaussian optimization without the rate constraint as the reference. Clear BD-Rate reductions are
observed for all tested settings, while $\beta=0.2$ achieves the lowest
BD-Rate of $-62.31\%$. Therefore, $\beta=0.2$ is adopted in the subsequent experiments.

\subsection{Combination with Gaussian Pruning}
\label{subsec:lightgaussian_generalization}

Gaussian pruning can be incorporated into the 3DGS compression pipeline using image codecs. 
To evaluate RDCompGS under this setting, we first adopt the pruning procedure in LightGaussian and optimize the Gaussian representation using RDCompGS.
In this process, pruning reduces the number of Gaussian primitives, while RDCompGS performs R-D optimization of the remaining Gaussian representation with respect to subsequent coding. 

As shown in Table~\ref{tab:lightgaussian_hevc}, the combination of RDCompGS with Gaussian pruning achieves consistent R-D improvements over LightGaussian on all three datasets. The BD-Rate reductions are 38.83\%, 67.19\%, and 54.54\% on Tanks \& Temples, Deep Blending, and Mip-NeRF 360,
respectively, corresponding to an average reduction of 53.52\%. 
Across different QPs, RDCompGS substantially reduces the compressed size while maintaining comparable or slightly higher PSNR and SSIM, with LPIPS remaining generally comparable to that of LightGaussian.
These results demonstrate that RDCompGS remains effective when combined with Gaussian pruning, providing additional R-D gains for the pruned
Gaussian representation.

\begin{table*}[!t]
\centering
\caption{R-D performance comparison with LightGaussian under H.265/HEVC compression. Here, Ours denotes RDCompGS with the pruning strategy proposed in LightGaussian. Size is measured in MB and
PSNR in dB.}
\label{tab:lightgaussian_hevc}

\footnotesize
\setlength{\tabcolsep}{2.0pt}
\renewcommand{\arraystretch}{1.15}

\begin{tabular}{@{}l c
cccc @{\hspace{2pt}}
cccc @{\hspace{2pt}}
cccc @{\hspace{2pt}}
cccc @{\hspace{2pt}}
c@{}}
\toprule

\multirow{2}{*}{Dataset}
& \multirow{2}{*}{Method}
& \multicolumn{4}{c}{QP = 4}
& \multicolumn{4}{c}{QP = 10}
& \multicolumn{4}{c}{QP = 16}
& \multicolumn{4}{c}{QP = 22}
& \multirow{2}{*}{\begin{tabular}[c]{@{}c@{}}
BD-Rate\\
(\%) $\downarrow$
\end{tabular}}
\\

\cmidrule(lr){3-6}
\cmidrule(lr){7-10}
\cmidrule(lr){11-14}
\cmidrule(lr){15-18}

&
& Size & PSNR & SSIM & LPIPS
& Size & PSNR & SSIM & LPIPS
& Size & PSNR & SSIM & LPIPS
& Size & PSNR & SSIM & LPIPS
&
\\

\midrule

\multirow{2}{*}{\begin{tabular}[c]{@{}l@{}}
Tanks \& Temples\\[-1pt]
(Train)
\end{tabular}}
& LightGaussian
& 15.367 & 21.52 & 0.780 & 0.238
& 12.431 & 21.46 & 0.776 & 0.241
& 9.692  & 21.24 & 0.767 & 0.251
& 7.088  & 20.67 & 0.736 & 0.283
& \multirow{2}{*}{-38.83}
\\

& Ours
& 14.657 & 21.52 & 0.781 & 0.237
& 10.983 & 21.47 & 0.778 & 0.240
& 7.012  & 21.34 & 0.768 & 0.251
& 3.868  & 20.76 & 0.743 & 0.278
&
\\

\midrule

\multirow{2}{*}{Deep Blending}
& LightGaussian
& 36.044 & 29.76 & 0.898 & 0.256
& 29.350 & 29.72 & 0.897 & 0.257
& 23.069 & 29.57 & 0.894 & 0.262
& 17.042 & 29.06 & 0.883 & 0.277
& \multirow{2}{*}{-67.19}
\\

& Ours
& 25.096 & 29.95 & 0.902 & 0.254
& 13.073 & 29.86 & 0.901 & 0.257
& 7.861  & 29.63 & 0.897 & 0.265
& 5.731  & 28.96 & 0.885 & 0.285
&
\\

\midrule

\multirow{2}{*}{Mip-NeRF 360}
& LightGaussian
& 39.069 & 26.65 & 0.789 & 0.232
& 31.674 & 26.54 & 0.786 & 0.236
& 24.776 & 26.21 & 0.776 & 0.247
& 18.223 & 25.38 & 0.749 & 0.276
& \multirow{2}{*}{-54.54}
\\

& Ours
& 32.578 & 26.77 & 0.791 & 0.233
& 21.254 & 26.64 & 0.787 & 0.239
& 11.891 & 26.22 & 0.776 & 0.254
& 7.085  & 25.35 & 0.750 & 0.282
&
\\

\bottomrule
\end{tabular}

\end{table*}

\subsection{Ablation Studies}
\label{subsec:ablation_studies}

We conduct ablation studies to evaluate the two main components of RDCompGS, including differentiable rate estimation and the rate constraint adjustment model. 
Their effects are evaluated separately under
H.265/HEVC compression.

\subsubsection{Effectiveness of Differentiable Rate Estimation}
\label{subsubsec:rate_estimation_ablation}

To evaluate the effectiveness of differentiable rate estimation, we remove the estimated rate term from Gaussian optimization while keeping the other
settings unchanged. As shown in
Table~\ref{tab:rate_estimation_hevc_ablation}, incorporating rate estimation consistently improves the R-D performance across all three datasets, with
BD-Rate reductions exceeding 50\%. In particular, the reduction reaches 62.32\% on Deep Blending, while positive BD-PSNR gains are obtained
on all three datasets. These results demonstrate that differentiable rate estimation effectively connects Gaussian representation optimization with
subsequent coding, leading to substantial R-D improvements.

\begin{table}[!t]
\centering
\caption{Ablation of differentiable rate estimation under H.265/HEVC compression, using Gaussian optimization without the estimated rate term
as the reference.}
\label{tab:rate_estimation_hevc_ablation}

\footnotesize
\setlength{\tabcolsep}{5pt}
\renewcommand{\arraystretch}{1.10}

\begin{tabular}{@{}lcc@{}}
\toprule
Dataset
& BD-Rate (\%) $\downarrow$
& BD-PSNR (dB) $\uparrow$ \\
\midrule

Tanks \& Temples (Train)
& -53.28
& 0.63 \\

Deep Blending
& -62.32
& 0.88 \\

Mip-NeRF 360
& -56.09
& 1.18 \\

\bottomrule
\end{tabular}

\end{table}

\subsubsection{Effectiveness of the Rate Constraint Adjustment Model}
\label{subsubsec:rate_constraint_ablation}

We further evaluate the rate constraint adjustment model while retaining differentiable rate estimation. Specifically, a single rate constraint with $\lambda_{\mathrm{rate}}=1\times10^{-7}$ is used across all compression conditions instead of the adjustment model, while the remaining settings are kept unchanged.

As shown in Table~\ref{tab:rate_constraint_adjustment}, the rate constraint adjustment model
provides non-negative BD-PSNR gains on all three datasets and improves the
overall BD-Rate performance. Although the BD-Rate increases slightly by
0.77\% on the Train scene, reductions are obtained on Deep Blending and
Mip-NeRF 360, with the latter reaching 5.46\%. Averaged over the three
datasets, the adjustment model achieves a BD-Rate reduction of 1.76\%. These results indicate that the adjustment model provides a more appropriate rate constraint under different compression conditions, leading to a more favorable overall R-D trade-off than using a single rate constraint.

\begin{table}[!t]
\centering
\caption{Comparison of the rate constraint adjustment model with the use of a single rate constraint under H.265/HEVC compression, using the latter as the reference.}
\label{tab:rate_constraint_adjustment}
\footnotesize
\setlength{\tabcolsep}{5pt}
\renewcommand{\arraystretch}{1.10}

\begin{tabular}{@{}lcc@{}}
\toprule
Dataset
& BD-Rate (\%) $\downarrow$
& BD-PSNR (dB) $\uparrow$ \\
\midrule
Tanks \& Temples (Train) & 0.77  & 0.00 \\
Deep Blending    & -0.59 &  0.02  \\
Mip-NeRF 360     & -5.46 &  0.25  \\
\bottomrule
\end{tabular}

\end{table}

\section{Conclusion}
\label{sec:conclusion}

In this work, we revisit 3DGS compression with 2D representations from the perspective of R-D optimization, with the aim of better aligning Gaussian attribute optimization with subsequent coding.
We introduce RDCompGS, which combines rate estimation with the rate constraint adjustment model to balance compression and rendering quality under different coding conditions. 
Extensive experiments on three benchmark datasets demonstrate consistent R-D improvements across different codec settings, and RDCompGS remains effective when Gaussian pruning is incorporated into compression with image codecs.

Future work will further explore joint modeling of coding rate and compression distortion. Extending RDCompGS to dynamic scenes is also an important direction for future research.

\bibliographystyle{IEEEtran}
\bibliography{ref}

\end{document}